\documentclass[lettersize,journal]{IEEEtran}
\usepackage{amsmath,amsfonts}
\usepackage{algorithmic}
\usepackage{algorithm}
\usepackage{array}
\usepackage[caption=false,font=normalsize,labelfont=sf,textfont=sf]{subfig}
\usepackage{textcomp}
\usepackage{stfloats}
\usepackage{url}
\usepackage{verbatim}
\usepackage{graphicx}
\usepackage{cite}
\usepackage{dsfont}
\usepackage{amsmath}
\usepackage{amssymb}
\usepackage{booktabs}
\usepackage{multirow}

\begin{document}

\title{Learning-Assisted Congestion-Aware Route Scheduling for Semiconductor Fab Material Control Systems}

\author{Hao Yin*, Meiqi Tu*, Anbang Liu, Shaochong Lin, Max Z.J. Shen%
\thanks{
\noindent This work has been submitted to the IEEE for possible publication. Copyright may be transferred without notice, after which this version may no longer be accessible.\\
*Hao Yin and Meiqi Tu contribute equally, and are the co-first author of this paper.\\
Hao Yin is with Department of Data and Systems Engineering, The University of Hong Kong, and is also with The Research Center for Digital Production and Smart Logistics, Tsinghua University.(yinh24@mails.tsinghua.edu.cn)\\
Meiqi Tu is with Department of Data and Systems Engineering, The University of Hong Kong (tumeiqi24@connect.hku.hk)\\
Anbang Liu is with Department of Data and Systems Engineering, The University of Hong Kong (anbang@hku.hk)\\
Shaochong Lin is with Department of Data and Systems Engineering, The University of Hong Kong (shaoclin@hku.hk)\\
Max Z.J. Shen is with Department of Data and Systems Engineering, The University of Hong Kong (maxshen@hku.hk)
}
}

\markboth{Preprint}%
{Anonymous \MakeLowercase{\textit{et al.}}: Learning-Assisted Congestion-Aware Route Selection for
Semiconductor Fab Material Control Systems}

\maketitle

\begin{abstract}
Automated material handling systems in semiconductor fabs are operated by a material control system (MCS) that must schedule a relay route for every transport command online, before execution. This is a data-driven scheduling problem in which route cost is dominated in the upper tail by queueing at heterogeneous, partially observable relay equipment, so route selection requires estimating both delivery time and congestion risk at the decision moment. This paper proposes a transport-network-aware dynamic congestion representation (TN-DCR). Built on a static directed transport graph induced by historically observed relay segments, TN-DCR combines structural route priors, multi-window network-wide congestion context, route-level bottleneck exposure, and an inductive graph-aware route embedding, all constructed under a prediction-time-safety invariant that admits only information observed strictly before the prediction moment. The representation feeds separate queue- and transfer-time regressors and an ordinal multi-label classifier producing calibrated multi-threshold exceedance scores, with an empirical-Bayes stock-key residual correction reducing systematic queue-time underprediction. The predictions serve as costs in a risk-constrained route-scheduling rule that minimizes predicted delivery time subject to a bound on extreme-congestion probability, embedding the learned predictors within a lightweight operations-research decision model. In a controlled closed-loop evaluation, mean delivery time falls by 16.4\% and internal resource waiting time by 22.6\% while throughput remains essentially unchanged.
\end{abstract}


\noindent\textit{Note to Practitioners}---This paper is motivated by a practical difficulty in operating a semiconductor fab AMHS: whenever the MCS routes a carrier, it must commit to a relay path online, before execution, without knowing how long the carrier will wait at intermediate stockers, lifts, and overhead transport. Such waiting is the main source of long, unpredictable deliveries and of queue-time risk for time-sensitive lots, yet current practice routes by a fixed shortest-path rule that ignores the live congestion state, and high-fidelity simulation is too slow to consult per command. We provide a lightweight decision aid that runs as each command arrives: it estimates queueing time, transfer time, and the probability of severe congestion for each candidate route, then selects a route that keeps this probability below an operator-set tolerance while minimizing predicted delivery time. Two properties matter for deployment. First, every feature is computed strictly from information observable at the decision moment, so offline evaluation reflects production behavior, and the gradient-boosted models support low-overhead candidate scoring without specialized hardware. Second, the risk ranking is concentrated: reviewing only the small fraction of highest-risk commands captures most truly congested ones, enabling selective manual intervention. Two limitations apply. The models are trained on routes chosen by the incumbent dispatcher, so predictions for rarely used routes are less reliable; the framework flags low-support routes and falls back to the existing policy instead of extrapolating. The reported gains come from a calibrated closed-loop simulator, so a staged, monitored rollout is recommended. Adapting the method to a new fab mainly requires re-estimating the transport graph, congestion statistics, and risk thresholds from that fab's own logs.

\section{Introduction}

\begin{figure}
    \centering
    \includegraphics[width=1\linewidth]{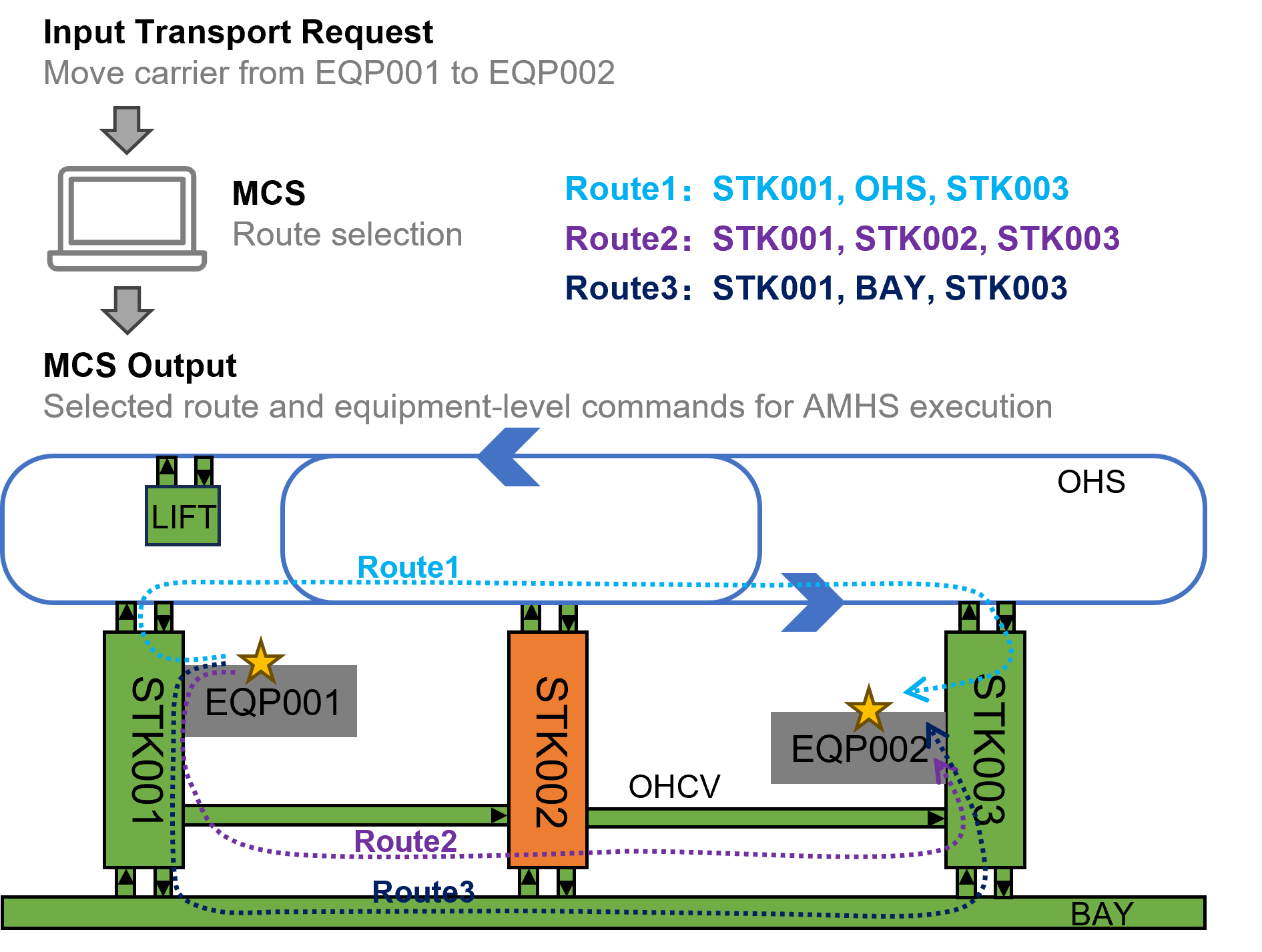}
    \caption{\textbf{Illustration of congestion-aware route selection in the MCS.} The MCS compares multiple feasible routes for transferring a carrier from EQP001 to EQP002. The orange STK002 denotes a congested stocker with many queued commands, which may increase the delivery delay of Route 2. The example shows why route evaluation should consider current congestion exposure at relay resources as well as static route length.}
    \label{fig:MCS_illustration}
\end{figure}

\IEEEPARstart{A}{s} advanced manufacturing becomes more automated and operates at higher tempos, Automated Material Handling Systems (AMHSs) have become critical to continuous production. This is especially true in semiconductor manufacturing, where fabs operate continuously and each carrier holds a batch of high-value in-process wafers or display panels. Material-handling delays directly increase cycle time and work-in-process, while prolonged queueing at intermediate storage can cause queue-time (Q-time) violations and threaten yield~\cite{roh2024kanban,zhang2025modeling}. As the logistics dispatching hub, the Material Control System (MCS) converts transport requests from the Manufacturing Execution System (MES) into executable commands and schedules a relay route for each command online. This is an AI-driven scheduling problem within data-driven manufacturing and logistics operations: it combines learned route-cost estimates with an operations-research decision rule under time-varying, partially observed congestion. Unlike the local controller of a single device, the MCS coordinates heterogeneous transport and relay resources. In semiconductor display fabs, these may include combinations of Stockers, Lifts, Bays, OHSs, and OHCVs (Fig.~\ref{fig:MCS_illustration}). The MCS must select suitable routes and resources for individual commands while maintaining efficient flow and limiting congestion.

Route selection is more complex than static shortest-path search. A fab AMHS may contain equipment from several vendors, with different interfaces and local control logic, while some internal device queues are only partly observable. A command may pass through several equipment types, so its delivery time depends on both route structure and the time-varying state of the associated resources. Congestion also propagates through the directed network: a congested stocker or lift can back-pressure the overhead transport it feeds and change the cost of neighboring bays. Transport times therefore have a pronounced upper tail, with a small fraction of severely delayed commands causing disproportionate operational disruption and Q-time risk. Before assigning a route, the MCS must estimate the expected delivery cost and congestion risk of each candidate using only information available at that moment. Command-level ETA and congestion-risk prediction are therefore essential for congestion-aware online scheduling.

Existing methods address parts of this problem but leave a gap between structural fidelity and online deployment. Discrete-event simulation captures AMHS topology, vehicle interactions, and local queues in detail, but is too expensive for low-latency evaluation of many candidate routes per command. Gradient-boosted decision trees efficiently handle heterogeneous engineered features in tabular industrial data and are attractive for command-level ETA prediction. However, the predictor alone does not directly encode the directed transport structure through which congestion spreads, so route topology, network-wide load, and route-local exposure are only indirectly represented. Graph-based road-network ETA models, including the production system in~\cite{derrow2021eta}, show the value of network structure. Yet they usually assume richly observed traffic and do not address prediction-time safety in an industrial MCS, where every feature must be constructed from information observable before a command is evaluated. Learning-based schedulers also show that predictive models deliver their value when embedded in an explicit decision model rather than replacing it~\cite{shi2025combination,tresca2024matheuristic}, but this perspective has received little attention in fab material handling.

We address this gap with a transport-network-aware dynamic congestion representation (TN-DCR), which provides the cost and risk terms for chance-constrained route scheduling in heterogeneous relay-based AMHSs. TN-DCR uses a static directed transport graph whose edges are induced by ordered node-pair traversals in historical routes. Dynamic information follows a strict prediction-time-safety invariant: every temporal feature uses only events and completed commands observed strictly before the command's prediction moment. This invariant is enforced across network-wide, route-local, and graph-neighborhood feature construction. It aligns the representation with the information boundary of an online MCS and prevents future or concurrent information from entering offline predictions.

Under this invariant, TN-DCR combines four types of information: structural priors describing the planned route and frequently traversed segments; multi-window, as-of context summarizing recent network-wide congestion immediately before prediction; route-level bottleneck exposure based on recent edge- and node-level activity along the candidate route; and a lightweight graph-aware route representation. The last component applies mean and max pooling to route nodes and their one-hop neighborhoods, followed by a fixed untrained projection. Unlike an end-to-end graph neural network, it has no learned message-passing parameters and can be recomputed independently of the downstream models. It therefore retains an inductive feature-construction mechanism for changing route compositions while avoiding the training and serving cost of a learned graph encoder. For route segments with limited historical support, explicit indicators allow unsupported candidates to be handled conservatively.

The same TN-DCR representation supports separate models for delivery time and congestion risk. Separate regressors estimate queueing and transfer times, from which total delivery time is reconstructed. An ordinal multi-label classifier models long-tail congestion by estimating exceedance probabilities at multiple nested thresholds from the same representation; the highest-threshold output is the most selective extreme-congestion indicator. An empirical-Bayes residual correction indexed by destination stock key further adjusts queue-time estimates. It reduces persistent destination-specific bias while shrinking sparsely observed stock keys toward the global model. Together, the regression and risk outputs provide the costs and constraints for a route-selection policy that limits the predicted extreme-congestion probability of the selected route. This follows chance-constrained routing under uncertainty~\cite{carpin2024solving,liu2024distributionally}, but the constraint uses an online, calibrated learned estimate rather than an external distributional assumption. The framework therefore lies between conventional tabular ETA models and graph- or event-aware congestion models such as~\cite{jin2023spatio}: it explicitly represents network structure and dynamic congestion while retaining lightweight predictors suitable for command-level online scoring.

The main contributions are as follows:
\begin{itemize}
    \item We formulate TN-DCR as a unified, prediction-time-safe representation for heterogeneous relay-based AMHSs. It combines structural route information, multi-window network-wide congestion, route-local bottleneck exposure, and graph-aware neighborhood information in a common feature space. All dynamic features use only information available before prediction.

    \item We develop a lightweight inductive graph representation using mean/max pooling over route nodes and their one-hop neighborhoods, followed by a fixed untrained projection. It requires no learned message passing and separates graph feature construction from downstream retraining, supporting changing route compositions and low-overhead online evaluation.

    \item We combine TN-DCR with separate queue- and transfer-time regressors, an ordinal multi-label risk model over several tail thresholds, and empirical-Bayes destination stock-key calibration for queue prediction. Their point and tail-risk estimates become the costs and constraints of a chance-constrained scheduling rule that minimizes predicted delivery time while bounding predicted extreme-congestion risk. This integrates learned predictors with a conventional operations-research decision model.

    \item We evaluate the framework on real transport-command data from a semiconductor display fab from both predictive and scheduling perspectives. The evaluation covers regressor selection under a shared feature, training, and postprocessing pipeline, multi-threshold risk assessment, ablations of dynamic and graph-aware features, stock-key calibration, and closed-loop comparison with static shortest-path routing. This links command-level prediction quality to downstream operational performance.
\end{itemize}

The remainder of the paper is organized as follows. Section~\ref{sec:related} reviews related work. Section~\ref{sec:formulation} defines the command-level prediction and route-selection problems and the prediction-time-safety requirement. Section~\ref{sec:method} presents TN-DCR, the prediction models, stock-key calibration, and route selection. Section~\ref{sec:results} reports the experiments, and Section~\ref{sec:conclusion} gives the limitations and future work.

\section{Related Work}
\label{sec:related}

We review four research threads related to our setting: scheduling and delivery-time estimation in fab material handling systems; learning-based transport scheduling and routing; network-aware travel-time and congestion prediction; and the use of learned predictors in optimization and risk-aware routing. These studies motivate a representation that captures network structure and dynamic congestion while supporting asynchronous, command-level MCS scheduling.

\subsection{Scheduling and Delivery-Time Estimation in Fab AMHS}
\label{sec:rel:amhs}

Semiconductor manufacturing is a demanding scheduling environment with strict timing constraints. Recent studies schedule cluster tools through deep reinforcement learning and adaptive search~\cite{kim2024scheduling}, control wafer-delay variability with Kanban feedback~\cite{roh2024kanban}, and model interacting timing requirements using dual-time Petri nets~\cite{zhang2025modeling}. They show that timing risk, rather than mean throughput alone, is central to fab scheduling. However, they focus on processing resources, whereas we schedule the relay path of each transport command.

AMHS performance has mainly been studied through queueing models and discrete-event simulation. Lin~\emph{et al.}~\cite{lin2003simulation} simulate connecting-transport layouts in 300-mm fabs, while Nazzal and McGinnis~\cite{nazzal2007expected} estimate response times for closed-loop multi-vehicle systems using an extended Markov-chain approximation. These methods support system-level analysis and design, but are not designed for command-specific routing under a rapidly changing network state. Repeated simulation is costly when many routes must be evaluated online, and analytical models rely on aggregate assumptions that do not capture the current route-dependent state. Liao and Wang~\cite{liao2004neural} estimate AMHS delivery time with separate neural predictors for intrabay loops and combine them into a path estimate. Although this matches the physical organization of a fab, it does not jointly model dependencies between successive segments and the current state of a directed transport graph.

These issues also arise beyond 300-mm wafer fabs. In the semiconductor display fab studied here, commands traverse heterogeneous relay equipment, intermediate storage creates queueing delay that affects Q-time and cycle-time risk, and a central controller coordinates movement over a structured network. We therefore study command-level relay-path scheduling whose cost depends on topology and current resource load. Unlike general warehouse or logistics routing, this setting also involves heterogeneous equipment, partially observed vendor-specific states, and yield-relevant timing constraints.

\subsection{Scheduling and Routing of Automated Transport in Manufacturing and Logistics}
\label{sec:rel:transport}

Related work also considers automated-vehicle dispatching and flow routing in manufacturing and logistics. Famularo~\emph{et al.}~\cite{famularo2024intelligent} coordinate autonomous vehicles through a multi-layer architecture that separates supervisory assignment from local motion control. Tresca~\emph{et al.}~\cite{tresca2024matheuristic} repair delivery routes as new information arrives, while related work balances retrieval flows by configuring automated storage modules~\cite{tresca2024matheuristics}. Learning-based methods include meta-reinforcement learning for dynamic flexible job shops~\cite{wu2025adaptive} and sparse-reward deep reinforcement learning for cloud manufacturing~\cite{wang2024hybrid}.

Our setting differs in two ways. First, these systems usually assume that tasks and resources are directly observable. A fab AMHS combines vendor-specific relay equipment whose internal queues are only partly reported, so the relevant state must be inferred from event logs. Second, their decisions mainly assign vehicles or workstations, while an MCS selects an ordered sequence of relay resources whose cost is strongly affected by intermediate queues. We therefore retain an explicit route-cost model and learn its congestion exposure.

\subsection{Network-Structure-Aware Travel-Time and Congestion Prediction}
\label{sec:rel:gnn-eta}

Route-time models can benefit from preserving connections among transport segments instead of treating a route as independent tabular attributes. Derrow-Pinion~\emph{et al.}~\cite{derrow2021eta} develop a production-scale graph-based ETA model, while DuETA~\cite{huang2022dueta} includes congestion propagation; inductive neighborhood aggregation supports generalization to unseen node combinations~\cite{hamilton2017inductive}. Other work predicts the network state itself. Diffusion-convolutional recurrent models~\cite{li2017diffusion} and adaptive-adjacency networks~\cite{wu2019graph} forecast grid-based flow or speed. Spatio-temporal graph fusion has also been used for industrial and environmental monitoring~\cite{qiao2024attention}. STGNPP instead models the occurrence and duration of congestion events as a marked point process on a road graph~\cite{jin2023spatio}.

A fab MCS differs in three respects. First, each command is evaluated asynchronously at its own decision time. All candidate-route features must use only information available then; a chronological train--test split alone does not prevent look-ahead leakage. We therefore impose prediction-time safety on the representation itself. Second, delay arises mainly from queues at discrete, capacity-limited relay resources whose internal states are only partly observed, rather than from free-flowing road traffic. Third, we do not predict a full future network snapshot or the next congestion event. For each command, we estimate candidate-route queue and transfer times and multiple upper-tail queue-risk probabilities. These levels distinguish routes with similar point estimates but different severe-delay exposure. We therefore use the directed AMHS topology to build graph-aware route features for lightweight predictors instead of training an end-to-end graph ETA model.

\subsection{Coupling Learned Predictors With Optimization-Based Decision Models}
\label{sec:rel:hybrid}

Gradient-boosted trees, including XGBoost~\cite{chen2016xgboost}, LightGBM~\cite{ke2017lightgbm}, and CatBoost~\cite{prokhorenkova2018catboost}, provide strong nonlinear modeling for heterogeneous tabular data with lightweight training and inference. A key industrial question is how to add graph structure without replacing this prediction stack with an end-to-end GNN. Ivanov and Prokhorenkova~\cite{ivanov2021boost} jointly train GBDT and GNN components for graphs with tabular node features. TN-DCR instead constructs a fixed-length graph-aware route representation through neighborhood aggregation and a fixed projection. Downstream GBDT models then use these features, separating graph feature construction from model fitting.

Another line of work combines learning with mathematical optimization. Shi~\emph{et al.}~\cite{shi2025combination} embed reinforcement learning in an optimization framework, allowing learned decisions to be evaluated against an explicit model rather than used as a black-box policy. Qin~\emph{et al.}~\cite{qin2026tackling} apply reinforcement learning to resource-sharing line balancing with a program-defined feasible region. Risk-aware routing offers related decision tools. Carpin~\cite{carpin2024solving} solves chance-constrained stochastic orienteering with Monte Carlo tree search, and Liu~\emph{et al.}~\cite{liu2024distributionally} study distributionally robust chance-constrained line planning. In these studies, the chance constraint comes from an external distributional model. In our setting, a calibrated ordinal classifier produces a command-specific exceedance probability from prediction-time-safe features. This learned risk estimate directly defines the scheduling rule's feasible set at the decision time.

Together, these threads reveal a gap. Fab scheduling studies focus on processing resources, while AMHS analytical and simulation methods do not support rapid, state-dependent evaluation of asynchronous commands. Road-network models use topology but assume richer observations and no command-specific information boundary. Hybrid learning--optimization methods often assume that uncertainty is known in advance. TN-DCR addresses this gap by constructing prediction-time-safe structural and dynamic features on a directed transport network, encoding them for lightweight tabular predictors, and using the resulting time and tail-risk estimates as the cost and chance constraint of congestion-aware route scheduling.

\section{Problem Formulation}
\label{sec:formulation}

We formulate congestion-aware routing in the material control system as two coupled tasks: edge-level prediction of delivery-time components and congestion risk using only information available at the route-decision moment, and path selection after aggregating these predictions over each admissible candidate. This section defines the transport system and path space, the prediction problem and its prediction-time-safety requirement, and the route-selection problem.

\subsection{System Model and Routing Objective}
\label{sec:form:setup}

For each transport command, the MCS selects a path through heterogeneous transport and relay resources before execution. The objective is to minimize predicted delivery time while limiting exposure to severe queueing under the time-varying network state.

The structural connectivity of the AMHS is represented by a static directed transport graph \(\mathcal{G}=(\mathcal{V},\mathcal{E})\), with \(\mathcal{E}\subseteq\mathcal{V}\times\mathcal{V}\), constructed from a historical reference corpus of logged transport paths. Each node \(v\in\mathcal{V}\) denotes a physical handling or relay location, and an edge \((u,v)\in\mathcal{E}\) indicates an observed directed transition. For each node \(v\), let \(\mathcal{N}^{\mathrm{out}}(v)=\{u:(v,u)\in\mathcal{E}\}\), \(\mathcal{N}^{\mathrm{in}}(v)=\{u:(u,v)\in\mathcal{E}\}\), and \(\mathcal{N}(v)=\mathcal{N}^{\mathrm{out}}(v)\cup\mathcal{N}^{\mathrm{in}}(v)\). The graph is fixed after construction from the reference-period data.

Observed connectivity alone does not determine path admissibility, since concatenating individually observed edges need not produce a path permitted by equipment-level control logic. For command \(c_i\), with origin \(o_i\in\mathcal{V}\) and destination \(d_i\in\mathcal{V}\), the admissible candidate-path set is \(\mathcal{P}_i=\mathcal{P}(o_i,d_i)\subseteq\{p:p\text{ is a loopless directed path from }o_i\text{ to }d_i\text{ in }\mathcal{G}\}\). It may be supplied by the lower-level routing system or generated subject to the path-feasibility and support constraints described in Section~\ref{sec:method:opt}.

The dynamic operating state is represented by the time-stamped event stream \(\mathcal{S}=\{s_j=(a_j,\tau_j,\boldsymbol{\zeta}_j)\}_j\), where \(\tau_j\in\mathbb{R}_{+}\) is the observation timestamp, \(a_j\in\mathcal{V}\cup\mathcal{E}\) identifies the associated node or edge, and \(\boldsymbol{\zeta}_j\in\mathbb{R}^{d_\zeta}\) contains its observed congestion-related quantities. These quantities include event counts and equipment states used to construct the dynamic features in Section~\ref{sec:method:pred}; histories of completed edge traversals provide an additional source of as-of information.

For command \(c_i\), let \(t_i\in\mathbb{R}_{+}\) denote the prediction moment after the command is received but before path assignment and transport execution. Its static attributes are denoted by \(\boldsymbol{x}_i\in\mathbb{R}^{d_x}\) and include origin and destination descriptors, carrier and priority information, and time-of-day variables.

A candidate path \(p=(v_{p,0},v_{p,1},\ldots,v_{p,L_p})\in\mathcal{P}_i\) satisfies \(v_{p,0}=o_i\) and \(v_{p,L_p}=d_i\). Its ordered directed-edge sequence is
\begin{equation}
\mathcal{E}(p)
=
\left(
e_{p,1},\ldots,e_{p,L_p}
\right),
\qquad
e_{p,k}
=
(v_{p,k-1},v_{p,k}).
\label{eq:path-edge-sequence}
\end{equation}
For edge \(e_{p,k}\), the edge-level input \(\boldsymbol{x}_{i,p,k}\) augments \(\boldsymbol{x}_i\) with the corresponding edge descriptors and the stock-key \(g_{i,p,k}\) used by the residual correction in Section~\ref{sec:method:head}.

The path-level times aggregate the corresponding edge-level quantities:
\begin{equation}
\begin{aligned}
y_i^{\mathrm{queue}}(p)
&=\sum_{k=1}^{L_p}y_{i,p,k}^{\mathrm{queue}}, \qquad
y_i^{\mathrm{trans}}(p)
=\sum_{k=1}^{L_p}y_{i,p,k}^{\mathrm{trans}},\\
y_i^{\mathrm{total}}(p)
&=y_i^{\mathrm{queue}}(p)+y_i^{\mathrm{trans}}(p).
\end{aligned}
\label{eq:delivery-decomp}
\end{equation}
Here, \(y_{i,p,k}^{\mathrm{queue}}\) is the delay before the transport resource required by \(e_{p,k}\) becomes available, and \(y_{i,p,k}^{\mathrm{trans}}\) is the subsequent transfer duration on that edge, both in seconds. After the selected path \(p_i^\star\) is executed, we omit its path argument and write \(y_i^{a}\equiv y_i^{a}(p_i^\star)\), \(a\in\{\mathrm{queue},\mathrm{trans},\mathrm{total}\}\).

Because the quantities in~\eqref{eq:delivery-decomp} are unavailable at \(t_i\), routing relies on edge-level time and risk predictions and their subsequent path-level aggregation.

\subsection{Prediction Problem}
\label{sec:form:targets}

The prediction problem is defined under an explicit information boundary. Let \(\mathcal{S}_{<t}=\{s_j\in\mathcal{S}:\tau_j<t\}\) contain the dynamic events observed strictly before \(t\), and let \(\mathcal{D}_{<t}\) contain the edge-level transport records completed before \(t\). The as-of information set is
\begin{equation}
\mathcal{I}_{t}
=
\left(
\mathcal{G},
\mathcal{S}_{<t},
\mathcal{D}_{<t}
\right).
\label{eq:infoset}
\end{equation}
All reference-period statistics and model parameters are frozen before evaluation, and every time-varying feature must respect the command-specific cutoff in~\eqref{eq:infoset}.

\begin{figure*}[t]
    \centering
    \includegraphics[width=\textwidth]{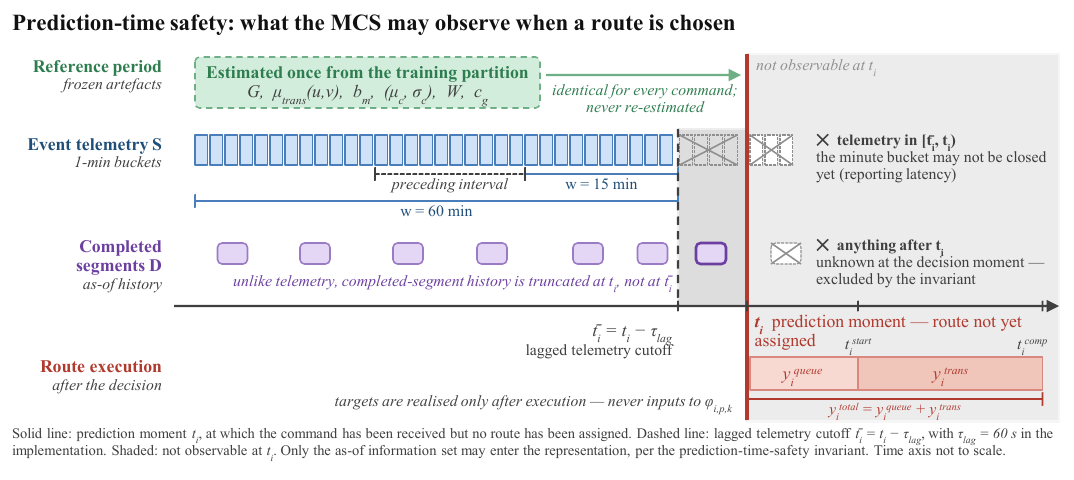}
    \caption{\textbf{Information boundary at the path-decision moment.}
    Minute-level telemetry is truncated at
    \(\bar{t}_i=t_i-\tau_{\mathrm{lag}}\), whereas completed-edge histories are truncated at \(t_i\). Both cutoffs precede path execution, and the shaded regions are unavailable at the decision moment.}
    \label{fig:timesafety}
\end{figure*}

For command \(c_i\), candidate path \(p\in\mathcal{P}_i\), and edge \(e_{p,k}\in\mathcal{E}(p)\), the feature representation satisfies
\begin{equation}
\boldsymbol{\phi}_{i,p,k}
=
f_{\phi}
\left(
\boldsymbol{x}_{i,p,k},
p,
\mathcal{I}_{t_i}
\right).
\label{eq:safety}
\end{equation}
This prediction-time-safety invariant excludes events and completed-edge outcomes unavailable at \(t_i\), thereby preventing look-ahead leakage. Fig.~\ref{fig:timesafety} illustrates the telemetry and completed-history cutoffs used in the implementation.

\subsubsection{Delivery-Time Targets}

The point-prediction task estimates queueing and transfer time separately for each edge \(e_{p,k}\), with base outputs \(\hat{y}_{i,p,k}^{\mathrm{queue}}\) and \(\hat{y}_{i,p,k}^{\mathrm{trans}}\). The stock-key calibration in Section~\ref{sec:method:head} is applied only to the queue estimate, producing \(\tilde{y}_{i,p,k}^{\mathrm{queue}}\). No separate total-time predictor is trained; path-level times are reconstructed from the edge-level outputs in Section~\ref{sec:form:opt}, consistent with~\eqref{eq:delivery-decomp}.

\subsubsection{Ordinal Congestion-Risk Targets}

Let \(\mathcal{D}^{\mathrm{ref}}\) denote the reference corpus, where record \(j\) is associated with edge \(e_j=(u_j,v_j)\). For \((u,v)\in\mathcal{E}\), define \(\mathcal{D}^{\mathrm{ref}}_{(u,v)}=\{j\in\mathcal{D}^{\mathrm{ref}}:(u_j,v_j)=(u,v)\}\) and \(N_{(u,v)}=|\mathcal{D}^{\mathrm{ref}}_{(u,v)}|\). For \(N_{(u,v)}>0\), the reference transfer-time scale is
\begin{equation}
\mu_{\mathrm{trans}}^{\mathrm{edge}}(u,v)
=
\frac{1}{N_{(u,v)}}
\sum_{j\in\mathcal{D}^{\mathrm{ref}}_{(u,v)}}
y_j^{\mathrm{trans}}.
\label{eq:edge-transfer-mean}
\end{equation}

For reference record \(j\), define \(\rho_j=y_j^{\mathrm{queue}}/\mu_{\mathrm{trans}}^{\mathrm{edge}}(u_j,v_j)\). For candidate edge \(e_{p,k}=(v_{p,k-1},v_{p,k})\), let \(\mu_{p,k}=\mu_{\mathrm{trans}}^{\mathrm{edge}}(v_{p,k-1},v_{p,k})\) and define
\begin{equation}
\rho_{i,p,k}
=
\frac{y_{i,p,k}^{\mathrm{queue}}}{\mu_{p,k}}.
\label{eq:queue-transfer-ratio}
\end{equation}
This ratio expresses queueing delay relative to the nominal transfer scale of the same edge, making congestion severity comparable across heterogeneous edges.

Let \(\mathcal{B}=\{b_1<\cdots<b_M\}\) contain thresholds estimated from the reference distribution of \(\rho_j\). For percentile level \(p_m\),
\begin{equation}
b_m
=
Q_{p_m}
\left(
\{\rho_j:j\in\mathcal{D}^{\mathrm{ref}}\}
\right),
\qquad
m=1,\ldots,M,
\label{eq:risk-threshold}
\end{equation}
and the corresponding edge-level ordinal label is
\begin{equation}
r_{i,p,k,m}
=
\mathds{1}
\left[
\rho_{i,p,k}\geq b_m
\right],
\qquad
m=1,\ldots,M.
\label{eq:risk-label}
\end{equation}
The labels are nested because \(b_1<\cdots<b_M\). In seconds, the edge-specific threshold is \(\theta_{p,k,m}=\mu_{p,k}b_m\), so \(r_{i,p,k,m}=\mathds{1}[y_{i,p,k}^{\mathrm{queue}}\geq\theta_{p,k,m}]\).

The shared TN-DCR representation feeds separate queue- and transfer-time regressors and an ordinal congestion-risk classifier, summarized by
\begin{equation}
F:
(\boldsymbol{x}_{i,p,k},p,\mathcal{I}_{t_i})
\mapsto
\left(
\tilde{y}_{i,p,k}^{\mathrm{queue}},
\hat{y}_{i,p,k}^{\mathrm{trans}},
\{\hat{r}_{i,p,k,m}\}_{m=1}^{M}
\right).
\label{eq:predictor}
\end{equation}
The edge identity is included in \(\boldsymbol{x}_{i,p,k}\), while \(p\) provides its candidate-path context; all outputs remain associated with \(e_{p,k}\). In particular, \(\hat{r}_{i,p,k,m}\approx P(\rho_{i,p,k}\geq b_m\mid\boldsymbol{x}_{i,p,k},p,\mathcal{I}_{t_i})\). Consistent with the nested labels, the predicted probabilities are constrained to satisfy \(\hat{r}_{i,p,k,1}\geq\cdots\geq\hat{r}_{i,p,k,M}\), with \(m=M\) representing the most selective edge-level risk. Their path-level aggregation is defined next.

\subsection{Route Optimization Problem}
\label{sec:form:opt}

For each candidate \(p\in\mathcal{P}_i\), the predictor in~\eqref{eq:predictor} is evaluated edge by edge while \(\mathcal{I}_{t_i}\) and the command-level attributes remain fixed. Using the edge scale \(\mu_{p,k}\) defined above, let \(\hat{\rho}_{i,p,k}=\tilde{y}_{i,p,k}^{\mathrm{queue}}/\mu_{p,k}\) denote the normalized predicted queueing ratio.

The edge-level time predictions are aggregated as \(\tilde{y}_{i}^{\mathrm{queue}}(p)=\sum_{k=1}^{L_p}\tilde{y}_{i,p,k}^{\mathrm{queue}}
\) and \(\hat{y}_{i}^{\mathrm{trans}}(p)=\sum_{k=1}^{L_p}\hat{y}_{i,p,k}^{\mathrm{trans}}\). Accordingly,
\begin{equation}
\begin{aligned}
\hat{y}_{i}^{\mathrm{total}}(p)
&=
\tilde{y}_{i}^{\mathrm{queue}}(p)
+
\hat{y}_{i}^{\mathrm{trans}}(p)\\
&=
\sum_{k=1}^{L_p}
\left(
\tilde{y}_{i,p,k}^{\mathrm{queue}}
+
\hat{y}_{i,p,k}^{\mathrm{trans}}
\right).
\end{aligned}
\label{eq:path-total-time}
\end{equation}
Thus, queueing severity is normalized at the edge level and converted back to seconds for the path-level delivery-time objective.

At ordinal level \(m\), the edge-level probabilities are aggregated into the path-level bottleneck-risk index
\begin{equation}
\hat{r}_{i,m}(p)
=
\max_{1\leq k\leq L_p}
\hat{r}_{i,p,k,m}.
\label{eq:path-risk-index}
\end{equation}
This aggregation requires no independence assumption, and \(\hat{r}_{i,m}(p)\leq\varepsilon\) requires every edge in \(p\) to satisfy the probability tolerance. However, \(\hat{r}_{i,m}(p)\) is a bottleneck-risk index rather than the calibrated probability that at least one edge on the path exceeds its threshold.

Using the highest ordinal level \(M\), define the surviving candidate set
\begin{equation}
\mathcal{P}_i^{(\varepsilon)}
=
\left\{
p\in\mathcal{P}_i:
\hat{r}_{i,M}(p)\leq\varepsilon
\right\}.
\label{eq:risk-filter}
\end{equation}
If \(\mathcal{P}_i^{(\varepsilon)}\neq\emptyset\), the selected path is \(p_i^{\star}=\arg\min_{p\in\mathcal{P}_i^{(\varepsilon)}}\hat{y}_i^{\mathrm{total}}(p)\). Otherwise, \(p_i^{\star}=\arg\min_{p\in\mathcal{P}_i}\hat{r}_{i,M}(p)\), with lower \(\hat{y}_i^{\mathrm{total}}(p)\) used to break ties.

This threshold-constrained rule is used throughout the paper. Because all candidate edges are evaluated using the same \(\mathcal{I}_{t_i}\), path comparison preserves the prediction-time-safety invariant in~\eqref{eq:safety}. Candidate generation, support filtering, and implementation of the selection rule are described in Section~\ref{sec:method:opt}.

\section{Methodology}
\label{sec:method}

This section instantiates the formulation in Section~\ref{sec:formulation} through TN-DCR feature construction, edge-level prediction and stock-key calibration, and candidate-path selection.

\subsection{Overview}
\label{sec:method:overview}

\begin{figure*}[t]
    \centering
    \includegraphics[width=\textwidth]{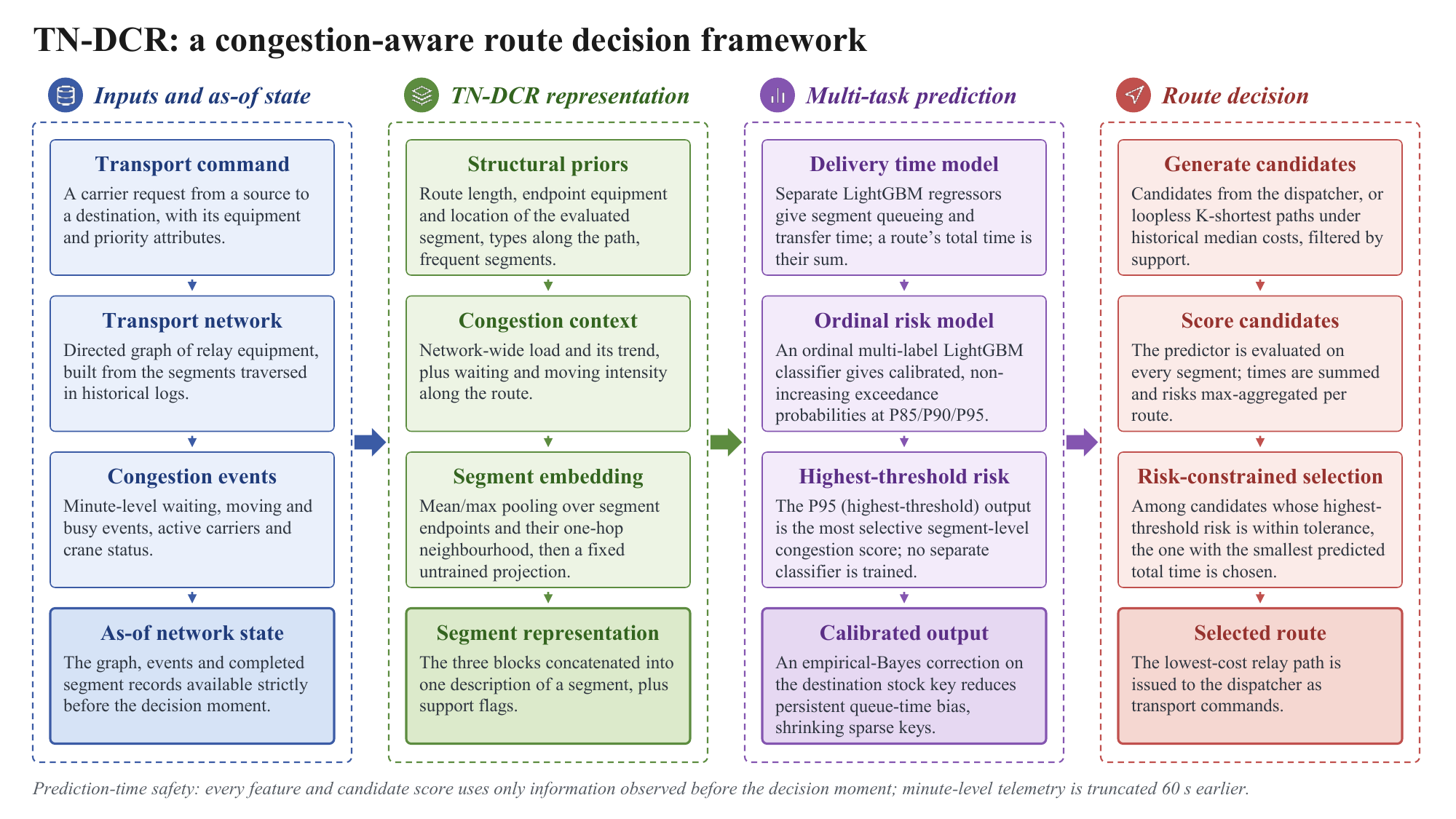}
    \caption{\textbf{Overview of the TN-DCR framework.}
    Structural, dynamic-context, and graph-aware features produce edge-level queue-time, transfer-time, and ordinal risk predictions. The corrected edge-level outputs are aggregated to evaluate and select among candidate paths.}
    \label{fig:framework}
\end{figure*}

Fig.~\ref{fig:framework} summarizes the framework. At prediction moment \(t_i\), TN-DCR combines command and edge attributes, candidate-path context, the reference graph, and as-of network information into an edge-level representation satisfying~\eqref{eq:safety}. This representation is shared by two LightGBM regressors for queue and transfer time and an ordinal multi-label LightGBM classifier, as formalized in~\eqref{eq:predictor}. The stock-key calibration is applied to the queue estimate, after which the edge-level outputs are aggregated according to~\eqref{eq:path-total-time} and~\eqref{eq:path-risk-index} for candidate-path selection.

Static graph artifacts and graph-derived feature construction are maintained independently of downstream LightGBM fitting, and the path-selection stage requires no additional model training.

\subsection{Prediction Model}
\label{sec:method:pred}

For each evaluated edge \(e_{p,k}\), TN-DCR decomposes the representation in~\eqref{eq:safety} as
\begin{equation}
\boldsymbol{\phi}_{i,p,k}
=
\left[
\boldsymbol{\phi}_{i,p,k}^{\mathrm{str}}
\Vert
\boldsymbol{\phi}_{i,p,k}^{\mathrm{ctx}}
\Vert
\boldsymbol{\phi}_{i,p,k}^{\mathrm{graph}}
\right]
\in
\mathbb{R}^{d_\phi},
\label{eq:tndcr-concat}
\end{equation}
where the three blocks encode structural information, dynamic congestion context, and graph-aware neighborhood information, respectively. Their modular construction supports the feature ablations reported in Section~\ref{sec:ablation}.

\subsubsection{Static Transport Graph and Structural Priors}
\label{sec:method:structure}

The graph \(\mathcal{G}\) defined in Section~\ref{sec:form:setup} is materialized from the reference partition after removing self-loops, invalid endpoints, and duplicate edges; traversal frequencies are estimated from the same data. It therefore represents historically observed directed transitions rather than the complete set of operationally admissible paths.

Incoming and outgoing neighbors are ordered by reference-period traversal frequency, and at most \(K_{\mathcal{N}}=48\) neighbors are retained for graph aggregation. Nodes and edges outside this fixed reference graph are treated as unsupported.

The structural block \(\boldsymbol{\phi}_{i,p,k}^{\mathrm{str}}\) includes the numbers of distinct path nodes and edges, endpoint equipment descriptors and location prefixes of \(e_{p,k}\), whether its endpoints share a base location, element and sub-element types along \(p\), and sparse indicators for frequently traversed edges. The latter retain \(K_{\mathrm{mac}}=60\) entries for the macro structural table and \(K_{\mathrm{cong}}=80\) for the congestion-oriented table. These features depend only on the command, evaluated edge, candidate path, and reference-period statistics.

\subsubsection{Dynamic Congestion Context}
\label{sec:method:context}

The context block \(\boldsymbol{\phi}_{i,p,k}^{\mathrm{ctx}}\) describes the network-wide and path-local state preceding the decision moment. The event stream is aggregated into one-minute buckets containing waiting- and moving-event counts, zone-busy activity, active-carrier counts, and crane-state indicators.

To exclude a potentially incomplete current bucket, telemetry uses
\begin{equation}
\bar{t}_i
=
t_i-\tau_{\mathrm{lag}},
\qquad
\tau_{\mathrm{lag}}=60\,\mathrm{s},
\label{eq:lagged-time}
\end{equation}
whereas completed-edge histories use the cutoff \(t_i\).

For each \(w\in\mathcal{W}=\{15,60\}\) minutes, the network-wide aggregate is \(\boldsymbol{\psi}_i^{\mathrm{net},w}=\sum_{\tau_j\in[\bar{t}_i-60w,\bar{t}_i)}\boldsymbol{\zeta}_j\). The half-open interval excludes observations at or after \(\bar{t}_i\), and comparison with the preceding interval of equal length captures short-term load changes.

Let \(\mathcal{V}(p)=\{v_{p,0},\ldots,v_{p,L_p}\}\), with \(\mathcal{E}(p)\) defined in~\eqref{eq:path-edge-sequence}. Over these path nodes and edges, the model computes event counts \(\mathrm{cnt}(v,w;\bar{t}_i)\) and intensity sums \(\mathrm{int}_{a}(v,w;\bar{t}_i)\), \(a\in\{\mathrm{wait},\mathrm{move},\mathrm{zone}\}\), and aggregates them by their sum, mean, maximum, active fraction, and concentration among the most active elements.

The waiting-to-moving intensity ratio is \(\eta_i^w(p)=\bigl(\sum_{v\in\mathcal{V}(p)}\mathrm{int}_{\mathrm{wait}}(v,w;\bar{t}_i)\bigr)\big/\bigl(\sum_{v\in\mathcal{V}(p)}\mathrm{int}_{\mathrm{move}}(v,w;\bar{t}_i)+\epsilon_0\bigr)\), where \(\epsilon_0>0\). It measures path-local waiting activity relative to movement and is distinct from the edge-level target ratio \(\rho_{i,p,k}\). All context features use observations preceding \(\bar{t}_i\).

\subsubsection{Graph-Aware Edge Representation}
\label{sec:method:gnn}

Unlike the path-wide context block, \(\boldsymbol{\phi}_{i,p,k}^{\mathrm{graph}}\) depends on the static graph connectivity around \(e_{p,k}\), independently of the remaining composition of \(p\).

For each node \(v\), temporally ordered cumulative event-count and intensity arrays are maintained. Binary search locates observations satisfying \(\tau_j<\bar{t}_i\), and cumulative differences over each \(w\in\mathcal{W}\) produce the windowed node state. Concatenating the two windows gives \(\boldsymbol{h}_v(\bar{t}_i)\in\mathbb{R}^{d_h}\).

For \(e_{p,k}=(v_{p,k-1},v_{p,k})\), define its endpoint set as \(\mathcal{V}(e_{p,k})=\{v_{p,k-1},v_{p,k}\}\) and its external neighborhood as \(\mathcal{N}(e_{p,k})=(\bigcup_{v\in\mathcal{V}(e_{p,k})}\mathcal{N}(v))\setminus\mathcal{V}(e_{p,k})\). At most \(K_{\mathcal{N}}\) neighbors are retained according to the reference-period ordering.

For any non-empty node set \(A\), define
\begin{equation}
    \begin{aligned}
    &\boldsymbol{m}_i^{A,\mu}
    =
    \frac{1}{|A|}
    \sum_{v\in A}
    \boldsymbol{h}_v(\bar{t}_i),\notag
    \\
    &\boldsymbol{m}_i^{A,\max}
    =
    \max_{v\in A}
    \boldsymbol{h}_v(\bar{t}_i).
    \end{aligned}
    \label{eq:pool}
\end{equation}

Both vectors are set to zero when \(A=\emptyset\). Applying~\eqref{eq:pool} to \(\mathcal{V}(e_{p,k})\) and \(\mathcal{N}(e_{p,k})\) gives their mean and maximum states. Their mean-state contrast is \(\boldsymbol{\chi}_{i,p,k}=\boldsymbol{m}_i^{\mathcal{V}(e_{p,k}),\mu}-\boldsymbol{m}_i^{\mathcal{N}(e_{p,k}),\mu}\).

The four pooled vectors and \(\boldsymbol{\chi}_{i,p,k}\) are concatenated into \(\boldsymbol{c}_{i,p,k}\in\mathbb{R}^{5d_h}\). To reduce skew, this vector is transformed as \(\tilde{\boldsymbol{c}}_{i,p,k}=\operatorname{sign}(\boldsymbol{c}_{i,p,k})\odot\log(1+|\boldsymbol{c}_{i,p,k}|)\) and standardized using reference-period statistics as \(\hat{\boldsymbol{c}}_{i,p,k}=(\tilde{\boldsymbol{c}}_{i,p,k}-\boldsymbol{\mu}_c)\oslash\boldsymbol{\sigma}_c\).

A fixed nonlinear projection produces \(\boldsymbol{\phi}_{i,p,k}^{\mathrm{emb}}=\tanh(\mathbf{W}^{\top}\hat{\boldsymbol{c}}_{i,p,k})\in\mathbb{R}^{d_e}\), where \(\mathbf{W}\in\mathbb{R}^{5d_h\times d_e}\), \(W_{ab}\sim\mathcal{N}(0,1/(5d_h))\), is initialized once and not optimized; \(d_e=16\). The shared projection provides a reproducible feature map for evaluated edges whose endpoint states and graph neighborhoods are represented, without implying reliable extrapolation to unsupported graph regions.

Two support diagnostics are appended to the embedding. With reference traversal count \(n^{\mathrm{ref}}(e_{p,k})\), the structural indicator is \(s_{p,k}^{\mathrm{str}}=\mathds{1}[n^{\mathrm{ref}}(e_{p,k})\geq n_{\min}]\). Let \(\mathcal{H}(e_{p,k})=\mathcal{V}(e_{p,k})\cup\mathcal{N}(e_{p,k})\); the dynamic support ratio is \(s_{i,p,k}^{\mathrm{dyn}}=|\{v\in\mathcal{H}(e_{p,k}):\boldsymbol{h}_v(\bar{t}_i)\text{ is available}\}|/|\mathcal{H}(e_{p,k})|\). 
The complete block is 
\[
\boldsymbol{\phi}_{i,p,k}^{\mathrm{graph}}=[\boldsymbol{\phi}_{i,p,k}^{\mathrm{emb}}\Vert s_{p,k}^{\mathrm{str}}\Vert s_{i,p,k}^{\mathrm{dyn}}].
\]

\begin{figure*}[t]
    \centering
    \includegraphics[width=\textwidth]{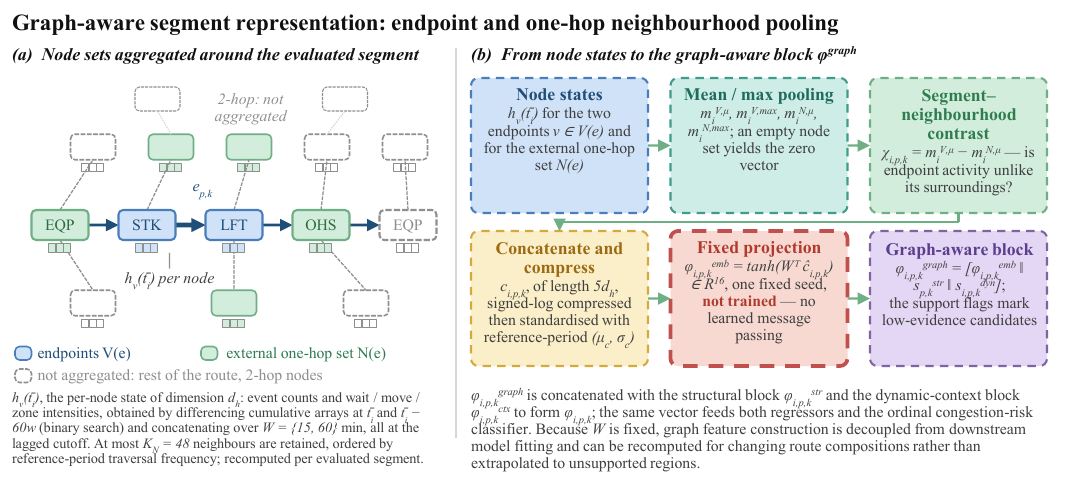}
    \caption{\textbf{Graph-aware edge representation.}
    Endpoint and neighboring-node states are mean/max pooled and contrasted, transformed by a fixed untrained projection, and concatenated with structural and dynamic support diagnostics.}
    \label{fig:graphrepr}
\end{figure*}

\subsubsection{Prediction Models and Stock-Key Residual Correction}
\label{sec:method:head}

Two separately fitted LightGBM regressors map \(\boldsymbol{\phi}_{i,p,k}\) to \(\hat{y}_{i,p,k}^{\mathrm{queue}}=f_{\mathrm{queue}}(\boldsymbol{\phi}_{i,p,k})\) and \(\hat{y}_{i,p,k}^{\mathrm{trans}}=f_{\mathrm{trans}}(\boldsymbol{\phi}_{i,p,k})\). No separate total-time regressor is trained.

Using the labels in~\eqref{eq:risk-label}, an ordinal multi-label LightGBM model produces \(\widehat{\boldsymbol{r}}_{i,p,k}=f_{\mathrm{risk}}(\boldsymbol{\phi}_{i,p,k})=(\hat{r}_{i,p,k,1},\ldots,\hat{r}_{i,p,k,M})\). The probabilities are calibrated on held-out reference-period data and constrained to satisfy \(\hat{r}_{i,p,k,1}\geq\cdots\geq\hat{r}_{i,p,k,M}\). The highest-threshold output is the most selective edge-level risk score; no separate extreme-congestion classifier is trained. The queue regressor and ordinal classifier are optimized independently, providing complementary point and exceedance-probability estimates.

Persistent stock-key queue residuals are handled using empirical-Bayes shrinkage. For a training edge record, let \(\hat{y}_{i,p,k}^{\mathrm{queue,OOF}}\) be its out-of-fold prediction and define \(\xi_{i,p,k}=y_{i,p,k}^{\mathrm{queue}}-\hat{y}_{i,p,k}^{\mathrm{queue,OOF}}\). 
For stock key \(g\), let \(\mathcal{J}_g=\{(i,p,k):g_{i,p,k}=g\}\), \(n_g=|\mathcal{J}_g|\), and \(\bar{\xi}_g=n_g^{-1}\sum_{(i,p,k)\in\mathcal{J}_g}\xi_{i,p,k}\). The correction is \(c_g=\frac{n_g}{n_g+\kappa}\bar{\xi}_g\), where \(\kappa>0\) shrinks estimates for sparsely observed keys toward zero.

The corrected edge-level queue estimate is \(\tilde{y}_{i,p,k}^{\mathrm{queue}}=\max(0,\hat{y}_{i,p,k}^{\mathrm{queue}}+c_{g_{i,p,k}})\), with \(c_{g_{i,p,k}}=0\) for unseen keys; the transfer estimate is used directly. The correction targets persistent stock-key bias rather than uniform improvement across all regression metrics.

\subsection{Route Optimization Algorithm}
\label{sec:method:opt}

The route-selection stage evaluates the admissible set \(\mathcal{P}_i=\mathcal{P}(o_i,d_i)\). When this set is supplied by the lower-level dispatcher, it is used directly; otherwise, a bounded set is generated by loopless \(K_p\)-shortest-path enumeration on \(\mathcal{G}\), using the reference-period median edge-level delivery times as base weights.

Candidates that violate the feasibility requirements of Section~\ref{sec:form:setup} or the minimum structural and dynamic support requirements are removed. If none remain, the command is passed to the incumbent routing fallback rather than evaluated outside the empirical support of the prediction model.

For each remaining path, the predictor in~\eqref{eq:predictor} is evaluated over its constituent edges using the same command attributes and \(\mathcal{I}_{t_i}\). The resulting edge-level outputs are aggregated using~\eqref{eq:path-total-time} and~\eqref{eq:path-risk-index}, after which the threshold-constrained rule in Section~\ref{sec:form:opt} is applied with risk tolerance \(\varepsilon\). Algorithm~\ref{alg:route} summarizes the procedure.

\begin{algorithm}[t]
\caption{Per-command TN-DCR path selection}
\label{alg:route}
\begin{algorithmic}[1]
\REQUIRE command \(c_i\) with origin \(o_i\), destination \(d_i\), and prediction moment \(t_i\); reference graph \(\mathcal{G}\); candidate source; trained edge-level predictors; support thresholds; risk tolerance \(\varepsilon\)
\ENSURE selected path \(p_i^{\star}\)

\STATE obtain admissible candidate set \(\mathcal{P}_i=\mathcal{P}(o_i,d_i)\)
\IF{the dispatcher does not provide \(\mathcal{P}_i\)}
    \STATE generate at most \(K_p\) loopless \(o_i\)-\(d_i\) paths using reference-period edge weights
\ENDIF

\STATE remove candidates violating feasibility or minimum-support requirements
\IF{\(\mathcal{P}_i=\emptyset\)}
    \STATE \textbf{return} incumbent routing fallback
\ENDIF

\FORALL{\(p\in\mathcal{P}_i\)}
    \FOR{\(k=1,\ldots,L_p\)}
        \STATE construct \(\boldsymbol{\phi}_{i,p,k}\) from \(\boldsymbol{x}_{i,p,k}\), \(p\), and \(\mathcal{I}_{t_i}\)
        \STATE evaluate the edge-level predictor in~\eqref{eq:predictor}
    \ENDFOR
    \STATE compute \(\hat{y}_i^{\mathrm{total}}(p)\) using~\eqref{eq:path-total-time}
    \STATE compute \(\{\hat{r}_{i,m}(p)\}_{m=1}^{M}\) using~\eqref{eq:path-risk-index}
\ENDFOR

\STATE form \(\mathcal{P}_i^{(\varepsilon)}\) using~\eqref{eq:risk-filter}
\IF{\(\mathcal{P}_i^{(\varepsilon)}\neq\emptyset\)}
    \STATE \(p_i^{\star}\leftarrow\arg\min_{p\in\mathcal{P}_i^{(\varepsilon)}}\hat{y}_i^{\mathrm{total}}(p)\)
\ELSE
    \STATE \(p_i^{\star}\leftarrow\arg\min_{p\in\mathcal{P}_i}\hat{r}_{i,M}(p)\)
    \STATE break ties using lower \(\hat{y}_i^{\mathrm{total}}(p)\)
\ENDIF

\RETURN \(p_i^{\star}\)
\end{algorithmic}
\end{algorithm}

Because the training edges belong to paths historically selected by the incumbent dispatcher, predictions for alternative candidates are observational rather than interventional. Support filtering limits extrapolation but does not eliminate selection bias; the policy therefore compares empirically supported candidates rather than estimating causal outcomes for arbitrary paths.

The policy is also myopic: it uses the network state at \(t_i\) without anticipating how future command reassignments will alter that state. Section~\ref{sec:closed_loop_results} evaluates this policy in a controlled closed-loop setting, while long-horizon fab-scale validation remains outside the scope of this study.

Internal generation considers at most \(K_p\) paths, with at most \(L_{\max}\) edge-level evaluations per path and \(K_{\mathcal{N}}\) retained graph neighbors per evaluation. Time-indexed histories are accessed by binary search, so candidate scoring is capped while history lookup grows only logarithmically with the event corpus. End-to-end serving latency remains an empirical deployment metric.

\section{Numerical Experiments}
\label{sec:results}

Using historical edge-level transport records from a semiconductor display manufacturing facility, we evaluate time regression, ordinal congestion-risk classification, feature and stock-key-correction ablations, and controlled closed-loop path selection.

\subsection{Experimental Setup}
\label{sec:experimental-setup}

\paragraph{Data and prediction protocol.}

The dataset contains \(51{,}700\) successful edge-level transport records collected from a production MCS in a semiconductor display fab. Each record is constructed at its command-specific prediction moment \(t_i\) and therefore obeys the information boundary in~\eqref{eq:safety}. 
The reference graph contains \(\lvert\mathcal{V}\rvert=762\) nodes and \(\lvert\mathcal{E}\rvert=1{,}549\) directed edges; approximately \(89\%\) of adjacent node pairs occur in only one direction. Queue time is strongly right-skewed, with mean \(63.3\,\mathrm{s}\), median \(9\,\mathrm{s}\), P95 \(285\,\mathrm{s}\), and P99 \(790\,\mathrm{s}\). Transfer time is less skewed, with mean \(116.7\,\mathrm{s}\), median \(77\,\mathrm{s}\), and a P95 approximately four times its median. The earliest \(80\%\) of records (\(41{,}360\)) are used for training and the latest \(20\%\) (\(10{,}340\)) for testing. The graph, reference statistics, and all other data-derived quantities are estimated from the training period only.

\paragraph{Regression targets and models.}

Separate LightGBM regressors predict edge-level queue and transfer times using the shared representation \(\boldsymbol{\phi}_{i,p,k}\) in~\eqref{eq:tndcr-concat}~\cite{ke2017lightgbm}. The stock-key calibration in Section~\ref{sec:method:head} is applied only to the queue output, while the transfer output is used directly. Edge-level total time is reconstructed from the two components; no separate total-time regressor is trained.

\paragraph{Ordinal congestion-risk labels.}

The edge-level ordinal targets follow the normalized ratio in~\eqref{eq:queue-transfer-ratio}, with transfer scales estimated from the training period using~\eqref{eq:edge-transfer-mean}. We set \(p_m\in\{0.85,0.90,0.95\}\) in~\eqref{eq:risk-threshold}, producing the P85, P90, and P95 labels in~\eqref{eq:risk-label}. Because these thresholds are estimated from the training distribution, their test prevalences need not equal exactly \(15\%\), \(10\%\), and \(5\%\).

The ordinal multi-label LightGBM classifier in Section~\ref{sec:method:head} uses the same TN-DCR representation and produces calibrated edge-level probabilities constrained to be non-increasing with congestion severity. These probabilities are evaluated as edge-level ranking scores and aggregated by~\eqref{eq:path-risk-index} for path selection.

\paragraph{Input features and model configuration.}

All learned models consume the same structural, dynamic-context, and graph-aware blocks in~\eqref{eq:tndcr-concat}. Dynamic telemetry is truncated at \(\bar{t}_i\) according to~\eqref{eq:lagged-time}, while completed-edge histories are truncated at \(t_i\).

\paragraph{Candidate regressors and non-learned references.}

History mean and History median assign the corresponding training-period statistic to every test record. The learned candidates are Random Forest~\cite{breiman2001random}, ExtraTrees~\cite{geurts2006extremely}, CatBoost~\cite{prokhorenkova2018catboost}, XGBoost~\cite{chen2016xgboost}, and LightGBM. All use the same chronological split, TN-DCR representation, training and evaluation protocol, and stock-key calibration, so the comparison isolates the choice of regressor. Each learned candidate is evaluated with five fixed random seeds (\(42\), \(365\), \(1234\), \(3407\), and \(114514\)), and results are reported as mean \(\pm\) standard deviation; the two history references are deterministic. Feature contributions are evaluated separately in Section~\ref{sec:ablation}.

\paragraph{Evaluation metrics.}

All metrics are computed on the chronological test partition. Let \(j\in\{1,\ldots,n\}\) index an edge-level record, with target \(y_j\), nonnegative prediction \(\hat{y}_j\), signed error \(\delta_j=\hat{y}_j-y_j\), positive-target set \(\mathcal{J}_{+}=\{j:y_j>0\}\), and \(n_{+}=|\mathcal{J}_{+}|\). We report
\begin{align}
\mathrm{MAE}
&=
\frac{1}{n}
\sum_{j=1}^{n}
|\delta_j|,
\notag\\
\mathrm{RMSLE}
&=
\sqrt{
\frac{1}{n}
\sum_{j=1}^{n}
\left[
\log(1+\hat{y}_j)-\log(1+y_j)
\right]^2
},
\notag\\
\mathrm{Cov}@\tau
&=
\frac{1}{n_{+}}
\sum_{j\in\mathcal{J}_{+}}
\mathds{1}
\left[
\frac{|\delta_j|}{y_j}\leq\tau
\right],
\qquad
\tau\in\{0.25,0.50\}.
\notag
\end{align}
MAE is reported in seconds, RMSLE measures logarithmic error, and Cov@25 and Cov@50 are the percentages of positive-target records whose absolute relative errors are within \(25\%\) and \(50\%\). Records with \(y_j=0\) are excluded from coverage because their relative errors are undefined. Lower MAE and RMSLE and higher coverage indicate better performance. For the stock-key analysis, \(\mathrm{Bias}=n^{-1}\sum_{j=1}^{n}\delta_j\), where a negative value indicates systematic underprediction.

For each congestion level, we report prevalence, PR-AUC, ROC-AUC, P@1\%, and P@5\%. The latter two are the positive fractions among the top \(1\%\) and \(5\%\) of edge-level records ranked by the corresponding exceedance probability. PR-AUC emphasizes performance under class imbalance, while top-ranked precision measures the concentration of high-risk records within a limited intervention set.

\subsection{Prediction and Classification Results}
\label{sec:prediction-results}

\begin{table}[t]
\centering
\caption{Edge-level regressor selection over five random seeds.}
\label{tab:overall-regression}
\scriptsize
\setlength{\tabcolsep}{3pt}
\renewcommand{\arraystretch}{1.05}
\resizebox{\columnwidth}{!}{%
\begin{tabular}{@{}llrrrr@{}}
\toprule
Target & Method & MAE (s) & RMSLE & Cov@25 (\%) & Cov@50 (\%) \\
\midrule
\multirow{7}{*}{Queue}
 & History mean   & 74.8 \(\pm\) 0.0 & 2.362 \(\pm\) 0.000 & 14.4 \(\pm\) 0.0 & 29.9 \(\pm\) 0.0 \\
 & History median & 68.4 \(\pm\) 0.0 & 2.202 \(\pm\) 0.000 & 14.0 \(\pm\) 0.0 & 29.9 \(\pm\) 0.0 \\
 & Random Forest  & 61.9 \(\pm\) 0.2 & 2.268 \(\pm\) 0.006 & 21.7 \(\pm\) 0.2 & 40.9 \(\pm\) 0.1 \\
 & ExtraTrees     & 63.6 \(\pm\) 0.2 & 2.345 \(\pm\) 0.009 & 20.0 \(\pm\) 0.2 & 39.8 \(\pm\) 0.1 \\
 & CatBoost       & 51.1 \(\pm\) 0.4 & \textbf{1.917 \(\pm\) 0.015} & 20.5 \(\pm\) 0.3 & 40.9 \(\pm\) 0.1 \\
 & XGBoost        & 51.1 \(\pm\) 0.6 & 1.928 \(\pm\) 0.032 & 21.0 \(\pm\) 0.4 & 41.2 \(\pm\) 0.3 \\
\cmidrule(lr){2-6}
 & \textbf{LightGBM}
 & \textbf{49.6 \(\pm\) 0.2}
 & 1.920 \(\pm\) 0.013
 & \textbf{22.6 \(\pm\) 0.2}
 & \textbf{43.6 \(\pm\) 0.3} \\
\midrule
\multirow{7}{*}{Transfer}
 & History mean   & 48.8 \(\pm\) 0.0 & 0.485 \(\pm\) 0.000 & 46.1 \(\pm\) 0.0 & 70.6 \(\pm\) 0.0 \\
 & History median & 41.3 \(\pm\) 0.0 & 0.480 \(\pm\) 0.000 & 67.6 \(\pm\) 0.0 & 86.0 \(\pm\) 0.0 \\
 & Random Forest  & 47.0 \(\pm\) 0.3 & 0.452 \(\pm\) 0.003 & 49.5 \(\pm\) 0.3 & 74.0 \(\pm\) 0.4 \\
 & ExtraTrees     & 43.2 \(\pm\) 0.5 & 0.420 \(\pm\) 0.005 & 43.7 \(\pm\) 1.2 & 71.3 \(\pm\) 0.8 \\
 & CatBoost       & 28.1 \(\pm\) 0.1 & 0.323 \(\pm\) 0.001 & 76.9 \(\pm\) 0.4 & 92.8 \(\pm\) 0.3 \\
 & XGBoost        & 29.6 \(\pm\) 0.5 & 0.334 \(\pm\) 0.003 & 74.8 \(\pm\) 1.2 & 91.2 \(\pm\) 0.8 \\
\cmidrule(lr){2-6}
 & \textbf{LightGBM}
 & \textbf{27.4 \(\pm\) 0.1}
 & \textbf{0.313 \(\pm\) 0.001}
 & \textbf{77.9 \(\pm\) 0.5}
 & \textbf{93.2 \(\pm\) 0.2} \\
\midrule
\multirow{7}{*}{Total}
 & History mean   & 103.0 \(\pm\) 0.0 & 0.657 \(\pm\) 0.000 & 31.2 \(\pm\) 0.0 & 56.3 \(\pm\) 0.0 \\
 & History median &  96.3 \(\pm\) 0.0 & 0.643 \(\pm\) 0.000 & 35.7 \(\pm\) 0.0 & 63.6 \(\pm\) 0.0 \\
 & Random Forest  &  92.6 \(\pm\) 0.4 & 0.584 \(\pm\) 0.002 & 35.1 \(\pm\) 0.3 & 60.1 \(\pm\) 0.2 \\
 & ExtraTrees     &  90.3 \(\pm\) 0.4 & 0.589 \(\pm\) 0.002 & 33.2 \(\pm\) 0.4 & 55.9 \(\pm\) 0.2 \\
 & CatBoost       &  67.1 \(\pm\) 0.3 & 0.446 \(\pm\) 0.002 & 51.6 \(\pm\) 0.4 & 79.1 \(\pm\) 0.3 \\
 & XGBoost        &  68.3 \(\pm\) 0.5 & 0.454 \(\pm\) 0.003 & 50.8 \(\pm\) 0.6 & 78.7 \(\pm\) 0.7 \\
\cmidrule(lr){2-6}
 & \textbf{LightGBM}
 & \textbf{64.9 \(\pm\) 0.2}
 & \textbf{0.433 \(\pm\) 0.001}
 & \textbf{52.9 \(\pm\) 0.4}
 & \textbf{79.7 \(\pm\) 0.3} \\
\bottomrule
\end{tabular}%
}
\end{table}

\paragraph{Regression model selection.}

As shown in Table~\ref{tab:overall-regression}, LightGBM provides the strongest overall performance. For queue time, it achieves the lowest MAE (\(49.6\pm0.2\,\mathrm{s}\)) and the highest Cov@25 and Cov@50 (\(22.6\pm0.2\%\) and \(43.6\pm0.3\%\)). CatBoost has a marginally lower queue RMSLE (\(1.917\pm0.015\) versus \(1.920\pm0.013\)), but the difference is small relative to the run-to-run variation. LightGBM achieves the best results on all transfer- and reconstructed-total-time metrics, including MAEs of \(27.4\pm0.1\,\mathrm{s}\) and \(64.9\pm0.2\,\mathrm{s}\), respectively, and is therefore selected for both regressors.

\paragraph{Multi-threshold congestion-risk classification.}

\begin{table}[t]
\centering
\caption{Edge-level congestion-risk classification over five random seeds.}
\label{tab:risk-detection}
\scriptsize
\setlength{\tabcolsep}{2pt}
\renewcommand{\arraystretch}{1.05}
\resizebox{\columnwidth}{!}{%
\begin{tabular}{@{}lrrrrrr@{}}
\toprule
Level & \(b_m\) & Prev. (\%) & PR-AUC & ROC-AUC & P@1\% & P@5\% \\
\midrule
P85
& 1.051
& 17.6 \(\pm\) 0.0
& 0.722 \(\pm\) 0.004
& 0.928 \(\pm\) 0.000
& 0.940 \(\pm\) 0.026
& 0.892 \(\pm\) 0.011 \\
P90
& 1.527
& 11.8 \(\pm\) 0.0
& 0.630 \(\pm\) 0.004
& 0.929 \(\pm\) 0.000
& 0.913 \(\pm\) 0.025
& 0.782 \(\pm\) 0.006 \\
P95
& 2.669
& 5.9 \(\pm\) 0.0
& 0.581 \(\pm\) 0.004
& 0.947 \(\pm\) 0.001
& 0.862 \(\pm\) 0.016
& 0.602 \(\pm\) 0.002 \\
\bottomrule
\end{tabular}%
}
\end{table}

Table~\ref{tab:risk-detection} reports the edge-level ordinal results. Test prevalences differ from their nominal training quantiles because of the chronological distribution shift. Across P85, P90, and P95, PR-AUC is \(0.722\pm0.004\), \(0.630\pm0.004\), and \(0.581\pm0.004\), while ROC-AUC remains at least \(0.928\). At P95, \(60.2\pm0.2\%\) of records within the highest-risk \(5\%\) are positive, compared with an overall prevalence of \(5.9\%\). These results characterize edge-level ranking; path selection uses the bottleneck aggregation in~\eqref{eq:path-risk-index}.

\subsection{Ablation Study}
\label{sec:ablation}

\paragraph{Graph-aware and dynamic-context features.}

Table~\ref{tab:queue-feature-ablation} ablates \(\boldsymbol{\phi}_{i,p,k}^{\mathrm{graph}}\) and \(\boldsymbol{\phi}_{i,p,k}^{\mathrm{ctx}}\) while holding the remaining pipeline fixed.

\begin{table}[t]
\centering
\caption{Edge-level queue-time feature ablation over five random seeds.}
\label{tab:queue-feature-ablation}
\scriptsize
\setlength{\tabcolsep}{3pt}
\renewcommand{\arraystretch}{1.05}
\resizebox{\columnwidth}{!}{%
\begin{tabular}{@{}lrrrr@{}}
\toprule
Variant & MAE (s) & RMSLE & Cov@25 (\%) & Cov@50 (\%) \\
\midrule
\textbf{Full}
    & \(\mathbf{49.6 \pm 0.2}\)
    & \(\mathbf{1.920 \pm 0.013}\)
    & \(22.6 \pm 0.2\)
    & \(\mathbf{43.6 \pm 0.3}\) \\
w/o graph
    & \(52.1 \pm 0.1\)
    & \(1.949 \pm 0.006\)
    & \(21.0 \pm 0.1\)
    & \(42.2 \pm 0.3\) \\
w/o dynamic
    & \(51.1 \pm 0.6\)
    & \(1.971 \pm 0.037\)
    & \(\mathbf{22.9 \pm 0.1}\)
    & \(42.4 \pm 0.1\) \\
w/o graph \& dynamic
    & \(54.4 \pm 0.4\)
    & \(2.075 \pm 0.020\)
    & \(22.1 \pm 0.3\)
    & \(42.4 \pm\) 0.1 \\
\bottomrule
\end{tabular}%
}
\end{table}

The full configuration gives the lowest MAE and RMSLE and the highest Cov@50. Removing the graph-aware or dynamic-context block increases MAE from \(49.6\pm0.2\,\mathrm{s}\) to \(52.1\pm0.1\,\mathrm{s}\) or \(51.1\pm0.6\,\mathrm{s}\), respectively. Removing both increases MAE to \(54.4\pm0.4\,\mathrm{s}\) (\(9.7\%\)), raises RMSLE from \(1.920\pm0.013\) to \(2.075\pm0.020\), and lowers Cov@50 from \(43.6\pm0.3\%\) to \(42.4\pm0.1\%\). Although removing dynamic context slightly improves Cov@25 from \(22.6\pm0.2\%\) to \(22.9\pm0.1\%\), the full representation is strongest overall, and the larger joint degradation indicates complementary contributions from the two blocks.

\paragraph{Stock-key residual correction.}

Removing the correction changes Bias from \(-9.6\pm0.6\,\mathrm{s}\) to \(-17.8\pm0.6\,\mathrm{s}\). The correction therefore reduces the magnitude of systematic underprediction by \(8.2\,\mathrm{s}\), or \(46.1\%\), although the remaining negative Bias indicates residual underestimation.

\subsection{Closed-Loop Routing Results}
\label{sec:closed_loop_results}

In a controlled closed-loop environment, we further compare TN-DCR routing with static shortest-path selection, which is a strategy that the factory previously used in actual operation, and the experiment tests whether aggregated edge-level predictions produce operational differences.

\paragraph{Evaluation environment.}

Because historical logs contain outcomes only for paths selected by the incumbent dispatcher, counterfactual policies are evaluated in a finite-capacity discrete-event simulator fitted exclusively from the training partition. On \(\mathcal{G}\), each edge and relay node is modeled as an independent multi-server FCFS resource, with capacities calibrated so that the busiest resource has target utilization \(0.70\). A selected path reserves its constituent resources and thereby changes the queues experienced by subsequent commands. Service times, capacities, and congestion slowdowns are fitted from raw logs without using TN-DCR predictions, providing all policies with the same independent ground-truth process.

Inter-arrival gaps, batch sizes, hourly demand, origin--destination pairs, and template paths are sampled from fitted or empirical training-period distributions. Edge service times follow the best-fitting exponential, log-normal, or gamma distribution and are divided into edge-movement and receiving-node handoff components. Bounded load-dependent slowdown, neighbor spillover, and rare incidents produce heavy-tailed congestion. Each replication generates \(500\) commands and discards the first \(50\) as warm-up. Results are averaged over five independent replications, with every policy evaluated on identical scenarios under a paired design.

\paragraph{Delivery-time components in the simulator.}

The simulator follows~\eqref{eq:delivery-decomp} but uses a specific accounting convention in which waiting after acquisition of the first resource belongs to transfer time. For command \(c_i\), let \(t_i^{\mathrm{arr}}\), \(t_i^{\mathrm{start}}\), and \(t_i^{\mathrm{comp}}\) denote its arrival, first-edge service start, and completion times. With the selected path implicit,
\begin{equation}
y_i^{\mathrm{queue}}
=
t_i^{\mathrm{start}}-t_i^{\mathrm{arr}},
\qquad
y_i^{\mathrm{trans}}
=
t_i^{\mathrm{comp}}-t_i^{\mathrm{start}}.
\label{eq:sim-components}
\end{equation}
Thus, queue time is the pre-dispatch delay, whereas transfer time spans the selected path and includes intermediate resource waiting.
We additionally report
\begin{equation}
y_i^{\mathrm{intwait}}
=
\left(
\sum_{r\in\mathcal{U}(p_i^\star)}
w_{i,r}
\right)
-
y_i^{\mathrm{queue}},
\label{eq:sim-intwait}
\end{equation}
where \(\mathcal{U}(p_i^\star)\) is the set of traversed edge and node resources and \(w_{i,r}\) is the waiting time at resource \(r\). Hence \(y_i^{\mathrm{intwait}}\) isolates the intermediate waiting contained within \(y_i^{\mathrm{trans}}\).

\paragraph{Compared policies.}

All policies use the same candidate set \(\mathcal{P}_i\), generated by loopless \(K_p\)-shortest-path enumeration with \(K_p=5\). \emph{Static shortest path} minimizes the sum of fixed edge weights given by historical median service times and does not react to the live state. \emph{TN-DCR routing} applies Algorithm~\ref{alg:route} with \(\varepsilon=0.05\). Candidate detours are limited to \(1.15\times\) the shortest static cost to prevent unbounded path-length increases. Table~\ref{tab:closed_loop_routing} shows that TN-DCR routing reduces mean and P95 total time by \(16.40\%\) and \(10.33\%\), respectively. Mean queue, transfer, and internal waiting times decrease by \(8.2\%\), \(18.9\%\), and \(22.59\%\).

\begin{table}[t]
\centering
\caption{Closed-loop performance averaged over five paired replications.}
\label{tab:closed_loop_routing}
\footnotesize
\renewcommand{\arraystretch}{1.08}
\setlength{\tabcolsep}{4pt}
\begin{tabular}{@{}lrr@{}}
\toprule
Metric & Static shortest path & TN-DCR routing \\
\midrule
Mean total time (s)       &  877.56 &  733.61 \\
P95 total time (s)        & 2584.43 & 2317.45 \\
Mean queue time (s)       &  202.87 &  186.31 \\
Mean transfer time (s)    &  674.70 &  547.30 \\
Mean internal wait (s)    &  564.75 &  437.15 \\
Throughput (commands/h)   &  295.43 &  295.68 \\
\bottomrule
\end{tabular}
\end{table}

The internal-wait reduction (\(127.60\,\mathrm{s}\)) is nearly equal to the transfer-time reduction (\(127.40\,\mathrm{s}\)), indicating that most of the transfer improvement comes from avoiding intermediate contention rather than shortening the physical path. This interpretation is consistent with the detour cap and with~\eqref{eq:sim-components}--\eqref{eq:sim-intwait}. The smaller queue-time reduction reflects the stronger dependence of pre-dispatch delay on origin-resource availability.
Throughput remains nearly unchanged at \(295.43\) versus \(295.68\) commands per hour. The experiment therefore demonstrates lower delivery and waiting times without an observed throughput difference under this operating condition. Absolute simulated times exceed those in the offline dataset because the simulator maintains sustained utilization and a heavy-tailed service process, and the results should therefore be interpreted as relative policy comparisons under identical calibrated scenarios. 

\subsection{Managerial Insights}
\label{sec:results-summary}

The results provide three operational implications for MCS management.

First, the P95 ranking concentrates positive cases within a limited intervention set: \(60.2\pm0.2\%\) of the highest-scored \(5\%\) of edge-level test records are positive, compared with a \(5.9\%\) overall prevalence. This more than tenfold enrichment supports selective operator review, path reconsideration, or priority handling.

Second, the feature blocks and stock-key calibration serve distinct purposes. Removing both graph-aware and dynamic-context features increases queue-time MAE from \(49.6\pm0.2\,\mathrm{s}\) to \(54.4\pm0.4\,\mathrm{s}\), a \(9.7\%\) degradation, although the coverage metrics are not uniformly monotonic. The stock-key calibration instead reduces Bias magnitude from \(17.8\pm0.6\,\mathrm{s}\) to \(9.6\pm0.6\,\mathrm{s}\), or \(46.1\%\), and should therefore be viewed primarily as a bias-control mechanism.

Third, closed-loop routing reduces mean total time by \(16.40\%\), P95 total time by \(10.33\%\), and internal waiting by \(22.59\%\), with nearly unchanged throughput. The dominant mechanism is avoidance of intermediate resource waiting rather than only pre-dispatch queue reduction. In deployment, support diagnostics should trigger conservative fallback for low-evidence paths, while fab-scale operation remains a separate validation step.

\section{Conclusion}
\label{sec:conclusion}

This paper formulated congestion-aware AMHS routing as edge-level prediction followed by candidate-path selection. TN-DCR combines structural path priors, multi-window network and path-local congestion context, and graph-aware endpoint-neighborhood information under a prediction-time-safe construction. Separate edge-level queue- and transfer-time LightGBM regressors and an ordinal risk classifier share this representation, while an empirical-Bayes stock-key calibration mitigates persistent queue bias. The edge outputs are aggregated into path-level time and bottleneck-risk estimates for risk-constrained selection. Graph information is encoded through endpoint/neighborhood pooling and a fixed nonlinear projection rather than an end-to-end trained graph encoder.

Experiments on \(51{,}700\) edge-level records identify LightGBM as the most consistent regressor over five seeds, with queue-, transfer-, and reconstructed-total-time MAEs of \(49.6\pm0.2\,\mathrm{s}\), \(27.4\pm0.1\,\mathrm{s}\), and \(64.9\pm0.2\,\mathrm{s}\). At P95, \(60.2\pm0.2\%\) of the highest-risk \(5\%\) of test records are positive, compared with a \(5.9\%\) prevalence. Ablations support the joint value of graph-aware and dynamic-context features, while stock-key calibration reduces Bias magnitude by \(46.1\%\). In the controlled closed-loop evaluation, TN-DCR routing reduces mean total time, P95 total time, and internal waiting by \(16.40\%\), \(10.33\%\), and \(22.59\%\), respectively, with nearly unchanged throughput.

The predictors remain observational because training records arise from paths chosen by the incumbent dispatcher; support filtering limits but does not eliminate this selection bias. Moreover, the max-aggregated path risk is a bottleneck index rather than a calibrated probability of any-edge threshold exceedance, and the closed-loop evaluation covers only a limited scale, horizon, and load regime. Future work should investigate off-policy path evaluation, calibrated path-level and denser tail-risk modeling, fab-scale validation under topology and load changes, learned graph aggregation, and end-to-end serving latency.


\bibliographystyle{IEEEtran}
\bibliography{ref}

\vfill

\end{document}